\documentclass[10pt,twocolumn,letterpaper]{article}

\usepackage{iccv}
\usepackage{times}
\usepackage{epsfig}
\usepackage{graphicx}
\usepackage{amsmath}
\usepackage{amssymb}
\usepackage{booktabs}
\usepackage{multirow}
\usepackage{algorithm}
\usepackage{algorithmic}
\usepackage{listings}
\usepackage{hhline}
\usepackage{bbding}
\usepackage[inkscapelatex=false]{svg}
\usepackage{soul}
\usepackage{color, xcolor}
\usepackage{dutchcal}
\usepackage[accsupp]{axessibility} 

\definecolor{battleshipgrey}{rgb}{0.52, 0.52, 0.51}

\usepackage[pagebackref=true,breaklinks=true,letterpaper=true,colorlinks,bookmarks=false]{hyperref}

\iccvfinalcopy 

\def\iccvPaperID{2564} 

\ificcvfinal\fi

\begin{document}

\title{TALL: Thumbnail Layout for Deepfake Video Detection}

\author{\textbf{Yuting Xu$^{1,3}$\thanks{This work was done when she was a student in CRIPAC.}\;, Jian Liang$^{2,4}$, Gengyun Jia$^{5}$, Ziming Yang$^{1,3}$, Yanhao Zhang$^{6}$, Ran He$^{2,4}\thanks{Corresponding author.}$}\\ 
$^{1}$ Institute of Information Engineering, Chinese Academy of Sciences\\
$^{2}$ CRIPAC \& MAIS, Institute of Automation, Chinese Academy of Sciences \\
$^{3}$ School of Cyber Security, UCAS 
$^{4}$ School of Artificial Intelligence, UCAS\\
$^{5}$ School of Communications and Information Engineering, NJUPT
$^{6}$ OPPO Research Institute\\
{\tt\small yuting.xu@cripac.ia.ac.cn, liangjian92@gmail.com, rhe@nlpr.ia.ac.cn}}

\maketitle
\ificcvfinal\thispagestyle{empty}\fi

\begin{abstract}
The growing threats of deepfakes to society and cybersecurity have raised enormous public concerns, and increasing efforts have been devoted to this critical topic of deepfake video detection.
Existing video methods achieve good performance but are computationally intensive.
This paper introduces a simple yet effective strategy named Thumbnail Layout (TALL), which transforms a video clip into a pre-defined layout to realize the preservation of spatial and temporal dependencies.
Specifically, consecutive frames are masked in a fixed position in each frame to improve generalization, then resized to sub-images and rearranged into a pre-defined layout as the thumbnail.
TALL is model-agnostic and extremely simple by only modifying a few lines of code.
Inspired by the success of vision transformers, we incorporate TALL into Swin Transformer, forming an efficient and effective method TALL-Swin.
Extensive experiments on intra-dataset and cross-dataset validate the validity and superiority of TALL and SOTA TALL-Swin.
TALL-Swin achieves 90.79$\%$ AUC on the challenging cross-dataset task, FaceForensics++ $\to$ Celeb-DF. The code is available at \url{https://github.com/rainy-xu/TALL4Deepfake}.
\end{abstract}

\section{Introduction}

\label{sec:intro}

Deepfakes generate and manipulate facial appearances to deceive viewers through generation techniques~\cite{zeng2022sketchedit,shi2022semanticstylegan}. 
With the remarkable success of generative adversarial networks~\cite{goodfellow2014,karras2019style}, deepfake products have become photo-realistic that humans can not distinguish.
These deepfake products~\cite{hsu2022dual,sun2022fenerf} may be misused for malicious purposes, leading to severe trust issues and security problems, such as financial fraud, identity theft, and celebrity impersonation~\cite{verdoliva2020media,mirsky2021creation}. 
The rapid development of social media exacerbates the abuse of deepfakes. 
Therefore, it is crucial to develop advanced detection methods to protect the data privacy of individual users. 

\begin{figure}[t]
\centering
\includegraphics[width=0.97\linewidth]{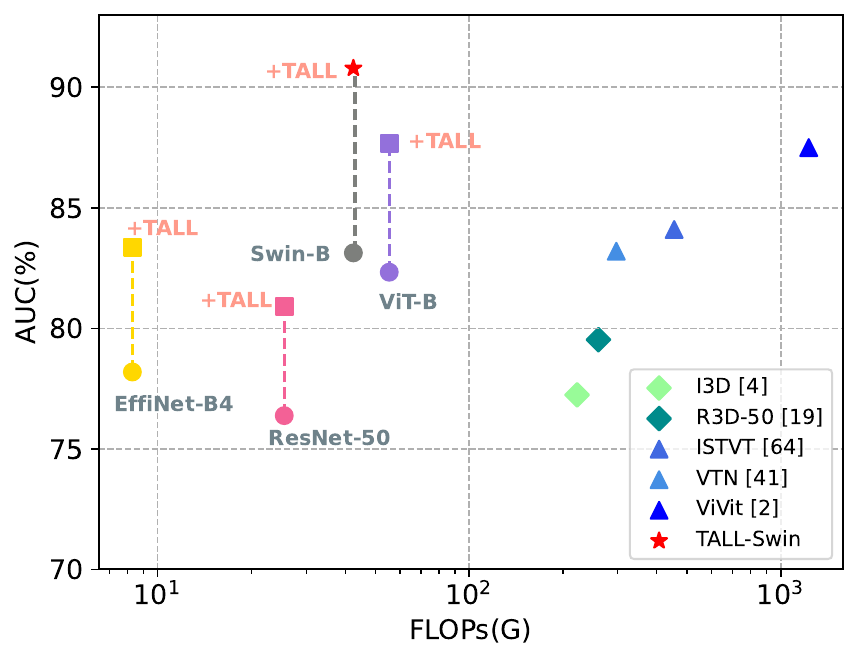}
\caption{\textbf{The AUC and FLOPs trade-off of different backbones.} Image-level backbones with TALL enjoy comparable accuracy-cost trade-offs with the 3DCNN and video transformer family on the unseen Celeb-DF dataset. All models with the same setting are trained on the FF++ (HQ)  dataset.
}

\label{fig:intro}
\end{figure}

Most previous image-based methods~\cite{huang2022fakelocator,zhu2021face} perform well on intra-dataset, but their generalizability needs to be improved. Recent research has focused on video-based methods to detect deepfake by modeling spatio-temporal dependencies. There are subtle spatio-temporal inconsistencies between frames since the deepfake algorithms are executed frame by frame.
The core of video-level approaches for deepfake detection is capturing inconsistencies through temporal modeling.
Existing deepfake video detection methods generally follow two directions. 
Some methods~\cite{gu2021spatiotemporal,gu2022delving} use two-branch networks or modules to learn spatial and temporal information separately and then fuse them. 
However, these two-branch approaches may fragment spatiotemporal cooperation and lead to subtle artifacts being neglected. 
Others directly use classic temporal models such as LSTM and 3D-CNNs. These methods are computationally intensive. 
The current rise of transformers for vision task backbones has prompted the emergence of corresponding deepfake detection methods. They are accompanied by significant computational complexity that makes them challenging to deploy and use, despite breakthroughs in performance. To enjoy benefits from both image and video methods, we are curious to see whether it is possible to append information about the temporal dimension to the image dimension.

This work develops a simple yet effective Thumbnail Layout (TALL) for deepfake detection by spatio-temporal modeling. TALL is computationally cheap and retains both temporal and spatial information. 
In detail, we use dense sampling to extract multiple clips in the video and then randomly select four consecutive frames in the video segment.
Subsequently, a block is masked at a fixed position in each frame.
Finally, the frames are resized as sub-image and sequentially rearranged into a pre-defined layout as a thumbnail, which has the same size as the clip frames. 
As shown in Figure~\ref{fig:intro}, TALL brings two advantages compared to the previous spatio-temporal modeling methods for deepfake detection: (1) TALL contains local and global contextual deepfake patterns. (2) TALL is a model-agnostic method for spatio-temporal modeling deepfake patterns at zero computation and zero parameters. 

Furthermore, we discover that the better temporal modeling capabilities backbone has, the better performance TALL achieves.
Based on the proposed TALL, we complement a baseline for video deepfake detection based on Swin Transformer~\cite{liu2021swin}, called TALL-Swin.
We validate TALL-Swin on four popular benchmark datasets, including FaceForensics++, Celeb-DF, DFDC, and DeeperForensics.  Our method gains a remarkable improvement over the state-of-the-art approaches.
The main contributions of our paper are summarized as follows:
\begin{itemize}
    \item We provide a new perspective for an efficient strategy for video deepfake detection called Thumbnail Layout (TALL), which incorporates both spatial-temporal dependencies, and allows the model to capture spatial-temporal inconsistencies.
    \item We propose a spatio-temporal modeling method called TALL-Swin, which efficiently captures the inconsistencies between deepfake video frames. 
    \item Extensive experiments demonstrate the validity of our proposed TALL and TALL-Swin. TALL-Swin outperforms previous methods in both intra-dataset and cross-dataset scenarios.
\end{itemize}

\section{Related Work}
\label{sec:rw}
\subsection{Image-Level Deepfake Detection}

Typically, existing deepfake detection methods fall into two categories: image-level and video-level methods. 
The image-level methods~\cite{jia2021inconsistency,fei2022learning} always exploit the artifacts of deepfake images in the spatial domain, such as discrepancies between local regions~\cite{nirkin2021deepfake, Yang_masked}, grid-like structure in frequency space~\cite{dong2022think}, and differences in global texture statistics~\cite{liu2020global} that provide specific clues to distinguish deepfakes from the real images. 
F3Net~\cite{qian2020} and FDFL~\cite{li2021fdfl} utilize the same pipeline that utilizes frequency-aware features and RGB information to capture the traces in different input spaces separately.
RFM~\cite{wang2021representative} and Multi-att~\cite{zhao2021multi} propose an attention-guided data augmentation mechanism to guide detectors to discover undetectable deepfake clues.
Face X-ray \cite{FaceXray} and PCL~\cite{self_cons} provide effective ways to outline the boundary of the forged face for detecting deepfakes. 
ICT~\cite{dong:2022} exploits an identity extraction module to detect identity inconsistency in the suspect image. 
Similarly, M2tr~\cite{wang:2022} detects local inconsistencies within frames at different spatial levels. 
Generally, image-level methods suffer over-fitting issues when a specific technique manipulates the images, and they ignore temporal information.
\subsection{Video-Level Deepfake Detection}
To improve the generalization of deepfake detectors, many studies generate diversity and generic deepfake data, while other studies capture the temporal incoherence of fake videos as generic clues. Some recent works  propose detecting temporal inconsistency using well-designed spatio-temporal neural networks, and others~\cite{gu2021spatiotemporal,gu2022delving} attempt to add modules to image models that capture temporal information.
STIL~\cite{gu2021spatiotemporal} formulates deepfake video detection as a spatial and temporal inconsistency learning process and integrates both spatial and temporal features in a unified 2D CNN framework. 
FTCN~\cite{zheng:2021} detects temporal-related artifacts instead of spatial artifacts to promote generalization. 
LipForensics~\cite{haliassos2021lips} is proposed to learn high-level semantic irregularity in mouth movement in the generated video. 
RealForensics~\cite{haliassos2022realforensics} uses auxiliary data sets during training in exchange for generalization at the cost of higher computational demands.
The video-based methods achieve strong generalization but suffer from large computational overhead.
To reduce computational costs, we propose TALL which gathers consecutive video frames into thumbnails for learning spatio-temporal consistency. 
\subsection{Deepfake Detection with Vision Transformer}
Recently, ViT~\cite{vit} has achieved impressive performance in computer vision tasks~\cite{ji2023masked,ji2022video,fan-iclr2022}. 
Many studies extend the ViT for deepfake detection~\cite{zhao:2022,wodajo:2021}. These methods achieve better performance compared to CNN-based models, but also sacrifice computational efficiency. 
A few works~\cite{wang:2022,zhao:2022} attempt to extend the transformer for deepfake detection due to the advent of the visual transformer (ViT) and the impressive ability to model long-range data, different from two-branch architectures that capture short-range and long-range temporal inconsistencies with a single-branch model. 
ICT~\cite{dong:2022} aims to detect identity consistency in deepfake video but may fail in detecting face reenactment and entire face synthesis results.
DFLL~\cite{khan:2021} extract the UV texture map to help the transformer to detect deepfakes, which may disrupt the continuity between video frames.
DFTD~\cite{khormali2022dfdt} leverages ViT to consider both global and local information but ignores the problem of excessive model arithmetic requirements.
Although the transformer-based approaches achieve promising performance, they are accompanied by significant computational complexity that makes them challenging to deploy and use, and the long-range dependencies may be insufficiently exploited in detection models. 
Swin Transformer~\cite{liu2021swin} produces a hierarchical feature representation and has linear computational complexity concerning input image size, which is suitable as a general-purpose backbone for various vision tasks. In this paper, we cooperate with Swin Transformer to form our robust and efficient method TALL-Swin.


\section{Method}
\label{sec:method}

TALL is a deepfake video detection strategy that transforms a video clip into an all-in-one thumbnail without the extra computational overhead. In the following sections, we begin with the motivation of TALL for deepfake detection in Section~\ref{sec:why}. Then we present the technical details of the TALL in Section~\ref{sec:TALL}. Finally, a generalizable Swin-TALL baseline is introduced to explore subtle artifacts in Section~\ref{sec:tall-swin}.

\subsection{Motivation} 
\label{sec:why}

While recent studies have attempted to address noticeable flaws through techniques like slight motion blurring and temporal consistency loss, subtle spatio-temporal artifacts still remain. These artifacts are important for detecting deepfakes, but they introduce two problems: 1) video-based models are less efficient, and 2) analyzing information over long distances may overlook local artifacts, which are critical for deepfake detection.
To address these challenges, we propose the TALL strategy, which naturally incorporates temporal information into image-level tasks without disrupting spatial information. This approach enables the image-level model to detect deepfakes in videos. Furthermore, we discovered that TALL provides even greater performance gains when combined with a powerful spatial model, resulting in the TALL-Swin.

\begin{figure}[t]
\centering
\includegraphics[width=0.96\linewidth]{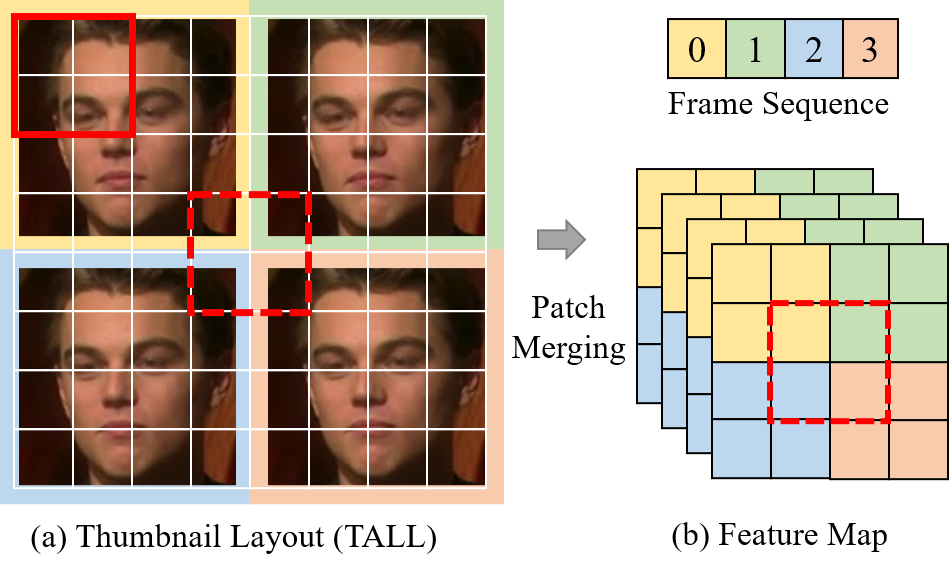}

\caption{Illustration of the TALL and shifted window process for computing self-attention in the TALL.}

\label{fig:TALL}
\end{figure}

In detail, TALL arranges consecutive frames in the temporal order in a compact 2$\times2$ layout, in line with the calculation theory of convolution and shifted window. 
TALL contains both spatial and temporal information so that model can learn both intra-frame artifacts and inter-frame inconsistency and obtains comparable performance to video-based methods.
Here we use the shifted window to explain TALL's mechanism. 
As illustrated in Figure~\ref{fig:TALL} (a), the model computes self-attention while accounting for spatial dependencies across sub-images (represented by the solid red box). When the window spans multiple sub-images (represented by the red dash box), the model is able to capture temporal inconsistencies between frames. Moreover, TALL leverages both local and global contexts of deepfake patterns to ensure robust modeling capabilities for short and long-range spatial dependencies. Compared to previous methods, we anticipate that TALL strikes a balance between speed and accuracy, sacrificing a little spatial information while preserving performance.
Based on the fact that attention-based models are better at handling contextual features and that the Swin-Transformer uses shifted windows to reduce computation and memory, we further complement
TALL-Swin baseline for video deepfake detection.

\subsection{Thumbnail Layout (TALL)} 
\label{sec:TALL}
Given a video $V\in\mathbb{R}^{{T}\times{C}\times{H}\times{W}}$, where  $T$ is the frame length of the video, $C$ is the number of channels, and ${H}\times{W}$ is the resolution of the frames.
Assuming each video contains $N$ clips, we divide a video into $N$ equal segments of length $T/N$ and then sample consecutive $t$ (set to 4 by default) frames from the segments at random locations to form one clip.
Then, the thumbnail $I$ is rearranged of sub-images (${C}\times{\frac{H}{\sqrt{t}}}\times{\frac{W}{\sqrt{t}}}$) that are resized from the above $t$ frames. To maximize the utility of TALL, we mask the organized $N$ square masks of the thumbnail. It is based on two core designs: 1) The position of the masks is random between different sub-images, which retains the advantages of the Cutout~\cite{devries2017improved} that encourages the network to focus more on complementary and less prominent features.
2) We fix the position of the mask within a clip to take advantage of the fact that most deepfake videos are frame-by-frame tampered with, thus forcing the model to detect inconsistencies between adjacent frames of the deepfake videos.
We do not allow the mask to appear on the seams of the thumbnail but allow for partial mask inclusion in the thumbnail. 
The detailed procedure of TALL is summarized in Algorithm~\ref{alg:code}.

\begin{algorithm}[t]
\caption{Pseudocode of TALL in a PyTorch-like style.}
\label{alg:code}

\definecolor{codeblue}{rgb}{0.25,0.5,0.5}
\lstset{
  backgroundcolor=\color{white},
  basicstyle=\fontsize{9pt}{9pt}\ttfamily\selectfont,
  columns=fullflexible,
  breaklines=true,
  captionpos=b,
  commentstyle=\fontsize{9pt}{9pt}\color{codeblue},
  keywordstyle=\fontsize{9pt}{9pt},
}
\begin{lstlisting}[language=python]
# x: one clip of video (T*C*H*W)
# T: frame number of clip
# C: channels; s: mask size
# d: length of frame included in the thumbnail
# r: rows of thumbnail
# x_tall: thumbnail image (224*224)

#TALL's augmentation strategy
h = np.random.randint(H)
w = np.random.randint(W)
#the mask position is fixed for each frame
m = np.ones((H, W))
h1 = np.clip(h - s // 2, 0, H)
h2 = np.clip(h + s // 2, 0, H)
w1 = np.clip(w - s// 2, 0, W)
w2 = np.clip(w + s // 2, 0, W)
m[h1: h2, w1: w2] = 0
m = torch.from_numpy(m)
m = mask.expand_as(x)
x = x * m
#TALL: generation of the thumbnail
x = x.view(-1,H,W).unsqueeze(0)
x = x.view((-1,C*d) + x_tall.size()[2:])
x = rearrange(x, `b (th tw c) h w 
     -> b c (th h) (tw w)`, th=r, c=C)
x_tall = interpolate(x, size=H)
\end{lstlisting}
\end{algorithm}

\subsection{TALL-Swin}
\label{sec:tall-swin}
To balance efficiency and model performance for spatio-temporal feature learning and to leverage the benefits of attention-based models, we enhanced a baseline deepfake detection model called TALL-Swin by incorporating the Swin Transformer~\cite{liu2021swin}.
Given the characteristics of TALL, we slightly modified the window size of Swin-B in TALL-Swin. 
We first enlarge the window size of the first three stages of the model so that the interaction between frames in the thumbnail becomes more frequent, forcing the model to learn more detailed spatio-temporal dependencies. 
Next, we set the window size of the last stage to be the same as the feature map size, enabling the window to perform global attention computations while TALL-Swin captures global spatial-temporal dependencies. As a result, the size of the last layer of the feature map became smaller, reducing the window size without introducing any additional computational overhead. Consequently, the window sizes for the four stages of TALL-Swin are $[14,14,14,7]$.
Note that the patch merging process makes TALL-Swin captures a more comprehensive range of dependencies through hierarchical representations, as shown in Figure~\ref{fig:TALL} (b). 

Given a video of length $T$, each frame contains $N$ patches, and the window contains $P$ patches.
To demonstrate the superiority of TALL-Swin in terms of computational consumption, we show below the computational complexity of the image-level transformer and video-level transformer, including ViT~\cite{vit}, Swin~\cite{liu2021swin}, ViViT~\cite{vivit}, and TALL-Swin respectively:
\begin{equation} 
\begin{array}{l}
{\rm{\Omega}_{ViT}} = 4TNC^2 + 2TN^2C, \\ [1mm]
{\rm{\Omega}_{Swin}} = 4TNC^2 + 2TPNC, \\ [1mm]
{\rm{\Omega}_{ViViT}} = 4TNC + 2T^2N^2C, \\ [1mm]
{\rm{\Omega}_{TALL-Swin}} = TNC^2 + \frac{1}{2}TPNC. 
\end{array}
\label{flops}
\end{equation}
TALL-Swin has the lowest computational complexity compared to image and video-level transformer methods.
Subsequent experiments will demonstrate that TALL-Swin maintains performance, albeit at the sacrifice of some spatial information.

The cross-entropy loss is employed to optimize the TALL-Swin, which is defined as:
\begin{equation}
\small
    \begin{aligned}{
        \mathcal{{L}}_{CE} = - \frac{1}{n} \sum_{i=1}^{n}{y_i \log{\mathcal{F}(x_i)}
        + (1-y_i) (\log{(1-  \mathcal{F}(x_i)})},
    }
    \end{aligned}
\end{equation}

where $x_i$ indicates input clip,  $y_i$ denotes the label of clip, $n$ is the number of clip, $\mathcal{F}$ is TALL-Swin.


\section{Experiments}
\label{sec:exp}

\subsection{Setup}
\label{sec:exp1}
\textbf{Datasets.} Following previous works~\cite{haliassos2021lips,haliassos2022realforensics,zheng:2021}, we evaluate the TALL and TALL-Swin on four widely used datasets. \textbf{FaceForensics++}~\cite{FaceForensics} is a most-used benchmark on intra-dataset deepfake detection, consisting of 1,000 real videos and 4,000 fake videos in four different manipulations: DeepFake~\cite{deepfake-faceswap}, FaceSwap~\cite{faceswap}, Face2Face~\cite{thies2016face2face}, and NeuralTextures~\cite{Thies:2019}. Besides, FaceForensics++ contains multiple video qualities, \eg high quality (HQ), low quality (LQ) and RAW. \textbf{Celeb-DF (CDF)}~\cite{li2020celeb} is a popular benchmark on cross-dataset, which contains 5,693 deepfake videos generated from celebrities. The improved compositing process was used to improve the various visual artifacts presented in the video. Celeb-DF is also suitable for deepfake detection tasks with a reference set. \textbf{DFDC}~\cite{dfdc} is a large-scale benchmark developed for Deepfake Detection Challenge. This dataset includes 124k videos from 3,426 paid actors. The existing deepfake detection methods do perform not very well on DFDC due to their sophisticated deepfake techniques. \textbf{DeeperForensics (DFo)}~\cite{jiang2020deeperforensics} includes 60,000 videos with 17.6 million frames for deepfake detection, whose videos vary in identity, pose, expression, emotion, lighting conditions, and blend shape with high quality.

\begin{table}[ht]
\centering
\setlength\tabcolsep{1.2pt}
\begin{tabular}{lcccccc}
\toprule
Models   & Temp. & CDF   & DFDC  & FLOPs  & Params & PT \\ \midrule
I3D-RGB$^*$~\cite{carreira2017i3d}  & \checkmark     & 78.24 & 65.58 & 222.7G & 25M     & 1K     \\ 
R3D-50$^*$~\cite{hara2017learning}    & \checkmark    & 79.63 & \textbf{67.73} & 296.6G & 46M     & 1K     \\ \midrule
ResNet50$^*$~\cite{He_2016_CVPR} & $\times$      & 76.38 & 64.01 & 25.5G  & 21M     & 1K     \\ 
+TALL    & \checkmark     & 80.90   & 65.54    & 25.5G  & 21M     & 1K     \\ \midrule
EffNetB4$^*$~\cite{tan2019efficientnet} & $\times$      & 78.19 & 66.81 & 8.3G   & 19M     & 1K     \\
+TALL    & \checkmark     & \textbf{83.37}     & 67.15  & 8.3G   &  19M   & 1K     \\ \midrule\midrule
VTN~\cite{neimark2021video}      & \checkmark     & 83.20 & 73.50 & 296.6G &46M    & 21K    \\
VidTR~\cite{zhang2021vidtr}    & \checkmark     & 83.30 & 73.30 & 117G   & 93M     & 21K    \\
ViViT$^*$~\cite{vivit}    & \checkmark     & 86.96 & 74.61 & 628G   & 310M    & 21K    \\
ISTVT~\cite{zhao2023istvt}    & \checkmark     & 84.10 & 74.20 & 455.8G & -       & -        \\ \midrule
ViT-B$^*$~\cite{vit}    & $\times$      & 82.33 & 72.64 & 55.4G  & 84M     & 21K    \\
+TALL    & \checkmark     & 86.58    & 74.10     & 55.4G  &  84M    & 21K    \\ \midrule
Swin-B$^*$~\cite{liu2021swin}   & $\times$      & 83.13 & 73.01 & 47.5G  & 86M     & 21K    \\
\textbf{TALL-Swin}   & \checkmark     & \textbf{90.79} & \textbf{76.78} &  47.5G  & 86M     & 21K    \\ \bottomrule
\end{tabular}
\vspace{0.5em}
\caption{\textbf{Performance of different backbones.} TALL consistently improves the accuracy over different image-level models. We show the AUC, FLOPs, and number of parameters for each model on the cross-dataset scenario. All models are trained on FF++ (HQ). \checkmark~indicates the model enables temporal modeling. * indicates our implementation. PT indicates pre-train. 1K and 21K indicate the model pre-trained on ImageNet-1K and 21K respectively. The best results are \textbf{bold}.} 
\label{tab:backcbone}
\end{table}

 \textbf{Implementation Details.}
We use MTCNN to detect face for each frame in the deepfake videos, only extract the maximum area bounding box and add 30$\%$ face crop size from each side as in LipForensics~\cite{haliassos2021lips}. The ImageNet-21K pretrained Swin-B model is used as our backbone. Excluding ablation experiments, we sample 8 clips using dense sampling, each clip contains 4 frames. The size of the thumbnail is $224\times$224. 
Following Swin Transformer~\cite{liu2021swin}, Adam~\cite{kingma2014adam} optimization is used with a learning rate of 1.5e-5 and batch size of 4, using a cosine decay learning rate scheduler and 10 epochs of linear warm-up. 
We adopt Acc. (accuracy) and AUC (Area Under Receiver Operating Characteristic Curve) as the evaluation metrics for extensive experiments. To ensure a fair comparison, we calculate video-level predictions for the image-based method and average the predictions across the entire video (following previous works~\cite{haliassos2021lips,gu2022delving,Liu_2021_CVPR,zheng:2021}).  Note that results are directly cited from published papers if we follow the same setting.

\subsection{Scaling over Backbones}
To verify our assumption, we adopt several image-level backbones commonly used for deepfake detection for comparison with the video-level backbones. 
As shown in Table~\ref{tab:backcbone} above the double horizontal line,  
we first compare the accuracy and complexity of the CNN-based video and image backbones. 
Although I3D~\cite{carreira2017i3d} and R3D~\cite{hara2017learning} achieve better performance than vanilla ResNet50~\cite{He_2016_CVPR} and EfficientNet~\cite{tan2019efficientnet}, the computation costs are huge, such as R3D-50 with 296G FLOPs.
For ResNet and EfficientNet who added TALL, ResNet achieves better AUC both on CDF (76.38 VS 80.93) and DFDC (64.01 VS 65.54) datasets. EfficientNet achieves 5.18$\%$ better AUC on CDF.

The second section contains the video and image transformers. Compared to video transformers, the image-based ViT and Swin fail to achieve better performance due to the lack of temporal modeling. For example, ViViT achieves 86.96$\%$ AUC on CDF, which is 3.6$\%$ higher than Swin although ViViT with 13$\times$ more computation. By way of contrast, ViT+TALL achieves 86.58$\%$ AUC on CDF with 55.4G FLOPs, which is comparable to AUC with ViViT but with low computation. Accordingly, Swin's performance was significantly improved with the addition of TALL without computation increment.
On the other hand, TALL boosts higher performance on models with learned long-range dependencies. 
\eg, ResNet+TALL (+4.5$\%$ on CDF and +1.5$\%$ on DFDC) \vs Swin+TALL (+7.6$\%$ on CDF and +3.6$\%$ on DFDC).
These two section results demonstrate that TALL provides both spatial and temporal information and enables the model to learn spatial and temporal inconsistencies for video deepfake detection. 

\begin{table}[ht]
\centering
\setlength\tabcolsep{7pt}{
\begin{tabular}{lcccc}
\toprule
\multicolumn{1}{l}{\multirow{2}{*}{Methods\quad}} & \multicolumn{2}{c}{ FF++ (HQ)} & \multicolumn{2}{c}{FF++ (LQ)} \\ \cmidrule(l){2-5}
\multicolumn{1}{c}{} &  Acc.    &AUC    &  Acc.      &  AUC    \\ \midrule

MesoNet~\cite{a:2018}  & 83.10 & -      & 70.47   & -      \\
Xception~\cite{xception} & 95.73 & 96.30      & 86.86 & 89.30        \\
Face X-ray~\cite{FaceXray}  & -      & 87.35 & -        & 61.60 \\
Two-branch~\cite{masi2020two}  & 96.43 & 98.70 & 86.34   & 86.59 \\
Add-Net~\cite{zi2020add}  & 96.78 & 97.74 & 87.50   & 91.01 \\
F3-Net~\cite{qian2020}                & 97.52 & 98.10 & 90.43   & 90.43 \\
FDFL~\cite{li2021fdfl}                  & 96.69 & 99.30 & 89.00   & 92.40 \\
Multi-Att~\cite{zhao2021multi}  & 97.60 & 99.29 & 88.69   & 90.40 \\
RECCE~\cite{Cao_2022_CVPR} &97.06 &99.32 &91.03 &95.02 \\ 
LipForensics~\cite{haliassos2021lips} &98.80 &99.70 & 94.20 & ~\textbf{98.10} \\
\midrule\midrule
DFDT~\cite{khormali2022dfdt}               &98.18   & 99.26      & 92.67  & 94.48     \\
ADT~\cite{wang2022adt}               & 92.05   & 96.30       & 81.48  & 82.52      \\
ST-M2TR~\cite{wang:2022}     & -   & 99.42    & -  & 95.31     \\ \midrule
VTN~\cite{neimark2021video}  & 98.47 & -  &  94.02  & - \\
VidTR~\cite{zhang2021vidtr} &  97.42 & - &  92.12 & -  \\
ViViT$^*$~\cite{vivit} &   92.60  & - & 88.02 & -  \\
ISTVT~\cite{zhao2023istvt} & \textbf{99.00}  & -      &  \textbf{96.15}  & - \\
\textbf{TALL-Swin}        &  98.65 &  \textbf{99.87}    &   92.82   & 94.57    \\ \bottomrule
\end{tabular}}
\vspace{0.5em}
\caption{\textbf{Intra-dataset evaluations.} We report the
video-level Acc. ($\%$) and AUC ($\%$) on the FF++ dataset. HQ indicates high quality, and LQ indicates low quality. } 
\label{table:ffpp}
\end{table}

\subsection{Comparison with State-of-the-art Methods
}
\textbf{Intra-dataset evaluations.} Following ISTVT~\cite{zhao2023istvt}, we show the results of the FF++ dataset under both Low Quality (LQ) and High Quality (HQ) videos, and report comparisons against several advanced methods in Table~\ref{table:ffpp}. We can observe that advanced video-based transformers have better results than CNN-based methods. 
Compared to video-based transformer methods, TALL-Swin has comparable performance and lower consumption to the previous video transformer method with HQ settings.
However, TALL-Swin gets unsatisfactory results with the LQ setting. The LQ setting is obtained by severely compressing the videos. So the reason for the result may be that TALL scales the frame to a smaller size, causing more spatial information to be lost in the frame.  We will investigate the possibility of other designs to further improve performance in the LQ setting.

\begin{table}[ht]
\centering

\setlength\tabcolsep{3pt} 
\begin{tabular}{lcccccccccccccc}
\toprule
Method         & CDF   & DFDC  & FSh   & DFo   & Avg.      \\ 
\midrule
Xception~\cite{xception}       & 73.70 & 70.90 & 72.00 & 84.50 & 75.28    \\
CNN-aug~\cite{wang2020cnnaug}        & 75.60 & 72.10 & 65.70 & 74.40 & 71.95    \\
CNN-GRU~\cite{sabir2019gru}        & 69.80 & 68.90 & 80.80 & 74.10 & 73.40    \\
Patch-based~\cite{chai2020patch}    & 69.60 & 65.60 & 57.80 & 81.80 & 68.70    \\
Face X-Ray~\cite{FaceXray}     & 79.50 & 65.50 & 92.80 & 86.80 & 81.15    \\
Multi-Att~\cite{zhao2021multi}     & 75.70 & 68.10 & 66.00 & 77.70 & 71.88    \\
DSP-FWA~\cite{li2019DSP}        & 69.50 & 67.30 & 65.50 & 50.20 & 63.13    \\
LipForensics~\cite{haliassos2021lips}  & 82.40 & 73.50 & 97.10 & 97.60 & 87.65    \\
\midrule \midrule
FTCN~\cite{zheng:2021}          & 86.90 & 74.00 & 98.80 & 98.80 & 89.63 \\
RealForensics~\cite{haliassos2022realforensics} & 86.90 & 75.90 & \textbf{99.70} & 99.30 & 90.45   \\ 
DFDT~\cite{khormali2022dfdt}          & 88.30 & 76.10 & 97.80 & 96.90 & 89.70    \\

\midrule
VTN~\cite{neimark2021video} & 83.20 & 73.50 & 98.70 & 97.70 & 88.30 \\
VidTR~\cite{zhang2021vidtr}  & 83.50 & 73.30 & 98.00 & 97.90 & 88.10 \\
ViViT$^*$~\cite{vivit} &86.96   & 74.61 & 99.41 &99.19 & 90.05 \\
ISTVT~\cite{zhao2023istvt}  & 84.10 & 74.20 & 99.30 & 98.60 & 89.10 \\
\textbf{TALL-Swin}          & \textbf{90.79}      & \textbf{76.78}    & 99.67      & \textbf{99.62}      & \textbf{91.71}          \\
\bottomrule
\end{tabular}
\vspace{0.5em}
\caption{ \textbf{Generalization to unseen datasets.} We report the video-level AUC ($\%$) on four unseen datasets: Celeb-DF (CDF), DFDC, FaceShifter (FSh), and DeeperForensics (DFo).}
\label{table:cross}
\end{table}

\textbf{Generalization to unseen datasets.}
In addition to the intra-dataset comparisons, we also investigate the generalization ability of our method. Adhering to the deepfake video detection cross-dataset protocol~\cite{haliassos2021lips}, we train a model on FF++ (HQ) then test on Celeb-DF (CDF), DFDC, FaceShifter (FSh), and DeeperForensics (DFo) datasets. As shown in Table~\ref{table:cross}: 
(1) Video-based methods generally have better results than image-based methods, which shows that temporal information is helpful for the deepfake video detection task. For example, Lip outperforms Face X-ray's AUC by a wide margin.
In addition, most transformer-based models have higher performance than CNN-based models. For the transformer-based models, both achieved an average AUC of 88$\%$, while the best CNN-based video-level models only achieved 87$\%$.
(2) TALL-Swin achieves state-of-the-art results on Celeb-DF, DFDC, and DeeperForensics datasets, and also beats its competitors on Celeb-DF dataset by a large margin (3.8$\%$). 
The results demonstrate that TALL-Swin performs well when encountering unseen datasets with better generalization ability than previous video transformer methods.

\begin{figure}[ht]
\centering
\includegraphics[width=0.98\linewidth]{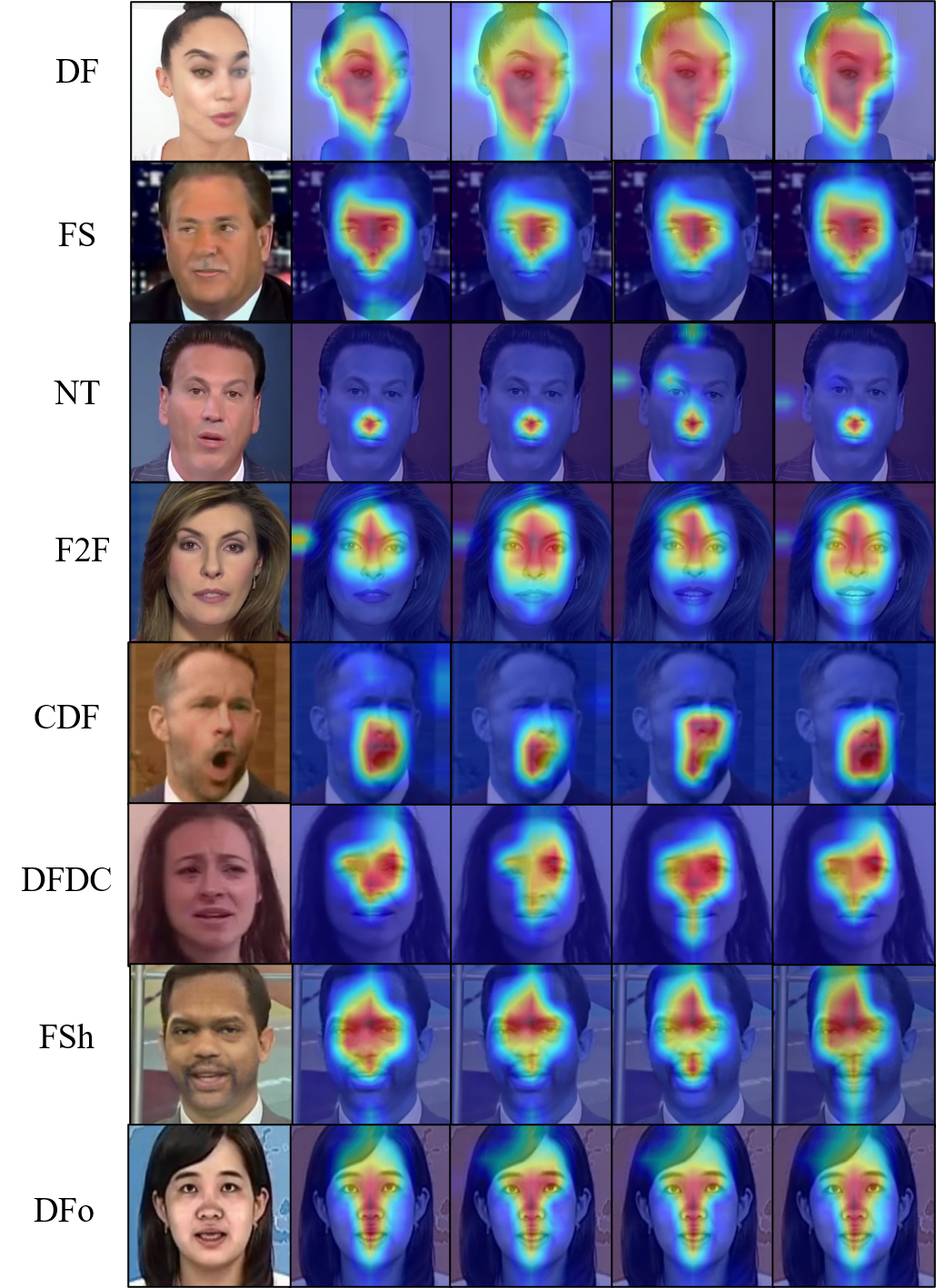}
\caption{\textbf{Saliency map visualization of TALL-Swin on different datasets.} The first four rows of samples are from the FF++ dataset, and the last four rows are from the unseen datasets. 
}
\label{fig:cam}
\end{figure}

\textbf{Analysis of saliency map visualization. } We adopt Grad-CAM~\cite{Selvaraju_2017_ICCV} to visualize where the TALL-Swin is paying its attention to the deepfake faces. In Figure~\ref{fig:cam}, we give the results on intra-dataset and cross-dataset scenarios. All models are trained on FF++ (HQ). 
It can be observed in the first four rows of Figure~\ref{fig:cam} that TALL-Swin captures method-specific artifacts. Note that the DF transfers the face region from a source video to a target, and the NT only modifies the facial expressions corresponding to the mouth region.
TALL-Swin corresponds to focus on the face region and the mouth region.
Furthermore, our model traces the more generalized artifacts that are independent of manipulation methods,~\eg, blending boundaries (CDF), and abnormal motions in the clip (DFDC, Fsh, Dfo). 

\begin{figure*}[htbp]
\centering
\includegraphics[width=1\textwidth]{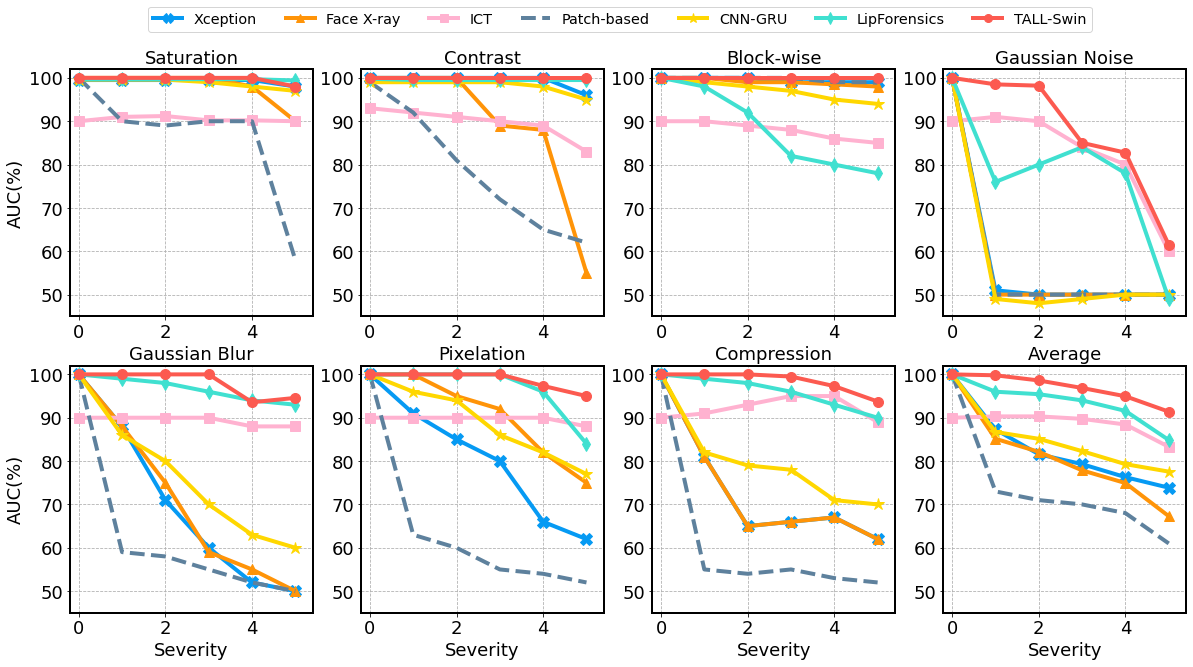}
\caption{\textbf{Robustness to various unseen corruptions.} We report the video-level AUC ($\%$) of our methods under five different levels of seven particular types of corruption. “Average'' denotes the mean across all corruptions at each severity level. Our TALL-Swin is more robust than previous methods for all corruptions. 
}
\label{fig:rob}
\end{figure*}

\begin{table*}[htbp]
\centering
\begin{tabular}{l c c c c c c c c c c c}
\toprule
Method  & {Clean} & Saturation & Contrast & Block & Noise & Blur & Pixel & Compress & Avg.  \\ 
\toprule
Xception (ICCV'19)~\cite{xception} & {99.8} & 99.3 & 98.6 & 99.7 & 53.8 & 60.2 & 74.2 & 62.1 & 78.3 \\
CNN-GRU (CVPRW'19)~\cite{sabir2019gru} & {99.9} & 99.0 & 98.8 & 97.9 & 47.9 & 71.5 & 86.5 & 74.5 & 82.3 \\
CNN-aug (CVPR'20)~\cite{wang2020cnnaug}  & {99.8} & 99.3 & 99.1 & 95.2 & 54.7 & 76.5 & 91.2 & 72.5 & 84.1  \\
Patch-based (ECCV'20)~\cite{chai2020patch} & {99.9} & 84.3 & 74.2 & 99.2 & 50.0 & 54.4 & 56.7 & 53.4 & 67.5   \\
Face X-ray (CVPR'20)~\cite{FaceXray}  & {99.8} & 97.6 & 88.5 & 99.1 & 49.8 & 63.8 & 88.6 & 55.2 & 77.5  \\
LipForensics (ICCV'21)~\cite{haliassos2021lips} & 99.9 & 99.9 & 99.6 & 87.4 & 73.8 & 96.1 & 95.6 & 95.6 & 92.5  \\
FTCN (ICCV'21)~\cite{zheng:2021}  & 99.4 & 99.4 & 96.7 & 97.1 & 53.1 & 95.8 & 98.2 & 86.4 & 89.5 \\
RealForensics (CVPR'22)~\cite{haliassos2022realforensics} & 99.8 & 99.8 & 99.6 & 98.9 & 79.7 & 95.3 & 98.4 & 97.6 & 95.6 \\
TALL-Swin $\mathcal{w/o}$ mask & \textbf{100.0} & \textbf{100.0} & \textbf{100.0} &99.8 & 83.5 & 97.3 & 98.4 & 97.9  &96.7\\
\textbf{TALL-Swin} &  \textbf{100.0} & \textbf{100.0} & \textbf{100.0} & \textbf{100.0} &\textbf{85.3} & \textbf{97.6} & \textbf{98.5} & \textbf{98.1} & \textbf{97.1}  \\
\bottomrule
\end{tabular}
\vspace{0.4em}
\caption{\textbf{Average robustness to unseen corruptions.} Average Video-level AUC ($\%$) across five intensity levels for each corruption type proposed in DFo~\cite{jiang2020deeperforensics}. “Avg'' indicates the mean across all corruptions and all levels.}
\label{tab:robustness}
\end{table*}

\textbf{Robustness to unseen perturbations.}
Deepfake detectors must be robust to common perturbations, given that video propagation on social media causes video compression, noise addition, etc. We also study the performance of robustness to unseen perturbations. Following RealForensics~\cite{haliassos2022realforensics}, the experiment applies seven unseen perturbations to fake videos at five intensity levels.
In Figure~\ref{fig:rob}, we show results of increasing the severity of each corruption. We can observe that other methods degrade dramatically as the perturbations become more severe. TALL-Swin still has a high performance. However, TALL-Swin degrades when the Gaussian noise reaches level five.
Table~\ref{tab:robustness} presents the average AUC across all intensity levels for corruption types. We observe that our method is significantly more robust to most perturbations than other methods. The good robustness may be from both the design of TALL and the proposed mask augmentation. The main reason may be the consecutive multi-frame input.  We empirically consider that the key to deepfake detection is local inconsistency, the continuous frame design has less redundant information, ensuring that the model finds locally important clues.

\begin{figure}[ht]
\centering
\includegraphics[width=0.8\linewidth]{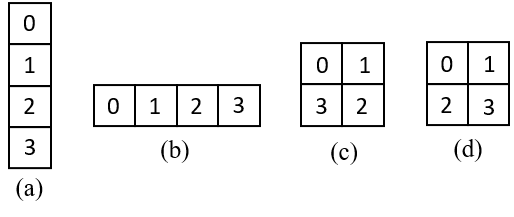}
\caption{\textbf{Illustration of different layout designs.}}
\label{fig:layout}
\end{figure}

\subsection{Ablation Study}

We perform the ablation study to analyze the effects of each component and hyper-parameter in TALL-Swin. 
All experiments are trained on FF++ (HQ) and tested on the CDF and DFDC datasets. 

\textbf{Effects of different layouts.} We train a TALL-Swin model on FF++(HQ) for each layout illustrated in Figure~\ref{fig:layout}, to analyze in which layout of the thumbnails the model learns the strongest generalization of the spatial-temporal dependence of the deepfake patterns.
As shown in Table~\ref{table1}, the model with a compact layout like Figure~\ref{fig:layout} (d) has good generalization ability on the unseen datasets.
A compact layout like Figure~\ref{fig:layout} (d) may help the model to learn the temporal dependence across frames because such a form provides the shortest distance between any two images.

\begin{table}[ht]
    \centering
    \small
    \begin{tabular}{cll}
    \toprule
    Layout & CDF & DFDC \\ \midrule
    Figure~\ref{fig:layout} (a)    &  85.52   &  70.02     \\
    Figure~\ref{fig:layout} (b)    & 84.93     & 73.57     \\
    Figure~\ref{fig:layout} (c)    & 86.66   &72.12      \\
    Figure~\ref{fig:layout} (d)    & \textbf{87.60} & \textbf{74.32}   \\ \bottomrule  
    \end{tabular}
    \vspace{0.5em}
    \caption{\textbf{Effects of different layouts.} All models here are trained without mask augmentation.}
    \label{table1}
\end{table}

\textbf{Study on the numbers of sub-image.} We use Swin-B as the baseline for this study to compare the effect of different thumbnail layout schemes on the model's generalization ability. 
Changing frames to thumbnails involves scaling, so we also investigate the impact of resizing and random cropping pre-processing on model performance. 
We set up four variants: resizing pre-process with ${4}\times{4}$ layout, ${3}\times{3}$ layout and ${2}\times{2}$ layout; random cropping pre-process with ${2}\times{2}$ layout.
As shown in Table~\ref{tb:frames}, the model performance degrades sharply when using ${4}\times{4}$ layout. This may be due to the small size of each sub-image that the spatial information is not captured well by the model. The result of ${3}\times{3}$ layout also slightly decreases. ${2}\times{2}$ layout with resizing pre-processing beats ${2}\times{2}$ layout with random crop. 
We also found that TALL-Swin achieves the best performance and the AUC score increases 3.2$\%$ compared to the baseline, suggesting that thumbnails in a ${2}\times{2}$ layout are more helpful to TALL-Swin than original frames.

\begin{table}[ht]
\centering

\setlength{\abovecaptionskip}{-0.4cm}
{
\setlength\tabcolsep{6pt}{
\begin{tabular}{ccccc}
\toprule
 Pre-process & Layout  & CDF & DFDC                 \\ \midrule
None  &   -  & 83.13    & 73.01     \\
Resize & ${4}\times{4}$  &80.18  &70.45   \\
Resize & ${3}\times{3}$  & 83.18 & 72.98  \\
Crop  & ${2}\times{2}$ & 78.55  & 73.30 \\
Resize & ${2}\times{2}$ &\textbf{87.60}  &\textbf{74.32}  \\ \bottomrule  
\end{tabular}
}}
\vspace{0.5em}
\caption{\textbf{Study on the numbers of sub-image.} All models here are trained without mask augmentation.}
\label{tb:frames}
\end{table}

\textbf{Effects of Sub-image's size.} 
We eliminate the scaling operation for sub-images to allow for more flexible layout settings. However, we've observed that when the number of sub-images grows at their original size, the performance improvements are only slight. Additionally, the computational complexity increases dramatically with the number of frames ( 4.3 times more than the TALL setting), as demonstrated in Table~\ref{tb:size}. To strike a balance between performance and computational complexity, we reduce the resolution of sub-images in TALL.

\begin{table}[ht]
\centering
\begin{tabular}{cccccc}
\toprule
Subimage-size & Layout & FLOPs  & CDF & DFDC            \\ \midrule
${224}\times{224}$ & ${3}\times{3}$ & 253G  & 88.69 & 75.98  \\
${224}\times{224}$ & ${2}\times{2}$ & 185G  & 88.15  & 75.01 \\
${112}\times{112}$ & ${2}\times{2}$ & 47.5G  &87.60  &74.32 \\ \bottomrule  
\end{tabular}
\vspace{0.5em}
\caption{\textbf{Effects of Sub-image's size.} All models here are trained without mask augmentation.
}
\label{tb:size}
\end{table}

\textbf{Study on absence and order of thumbnails.} 
In this case, we study the impact of missing the last sub-image and the last two sub-images on the model's performance. 
The first two rows of Table~\ref{table:order} show that all four sub-images contribute to the model performance. 
Besides, we set the order of the different thumbnails to evaluate the TALL-Swin. We consider three orders: forward, reverse, and random.
Forward order performs the best for three different orders. This may be because of the positional encoding of different frames in TALL-Swin.

\begin{table}[ht]
\centering

\setlength\tabcolsep{10pt}{
\begin{tabular}{ccc}
\toprule
Variants  & CDF & DFDC \\ \hline
0, 1, 2, -    &   86.46          &  69.51   \\ 
0, 1, -, -    &    84.22        &  69.09   \\ 
Random      &   85.85        &  70.30    \\
Reverse    &   86.65      &  72.37    \\
Forward    &   \textbf{87.60}      & \textbf{74.32} \\
\bottomrule  
\end{tabular}
}
\vspace{0.5em}
\caption{\textbf{Ablation study of absence and order of thumbnails.} All models here are trained without mask augmentation.}

\label{table:order}
\end{table}

\textbf{Effects of different orders on other backbones.} In order to prove that the phenomenon is not incidental, we also conduct experiments on ResNet50 and EfficientNet for three different orders as shown in Table~\ref{table:order2}. As expected, forward order outperforms reverse and random orders both on ResNet50 and EfficientNet, which indicates that TALL can learn the temporal dependency.

\begin{table}[ht]
\centering
\setlength\tabcolsep{6pt}{
\begin{tabular}{lcc}
\toprule
Variants  & CDF & DFDC  \\ \midrule
ResNet50+TALL     & 76.38  & 64.01       \\
Random    & 78.14    & 64.12   \\
Reverse   & 78.54    & 64.87    \\
Forward    & 80.90    & 65.54        \\  \midrule
EfficientNet+TALL    &78.19        & 66.81 \\
Random      & 81.01     & 66.13  \\
Reverse    &    81.66     &  66.69   \\
Forward    &83.37        &  67.15 \\ \bottomrule
\end{tabular}
}
\vspace{0.5em}
\caption{\textbf{Ablation studies of different orders of TALL.} }

\label{table:order2}
\end{table}

\textbf{Effectiveness of mask strategy.} In this work, TALL-Swin is trained on the FF++ (HQ) dataset without any data enhancement as the baseline except for Multi-scale Crop and Random Horizontal Flip.
To validate the effectiveness of the mask strategy, we compare our default baseline with different data augmentation strategies: 1) The Cutout~\cite{devries2017improved} on one sub-image; 2) The Cutout on four sub-images. 3) The combination of Mixup~\cite{zhang2017mixup} and Cutmix~\cite{yun2019cutmix} on four sub-images, as shown in Table~\ref{tb:cutout}.
The performance of a random Cutout~\cite{devries2017improved} on four sub-images is better than on one sub-image. 
Besides, the mask strategy leads to better performance than the well-known Cutout ($1.46\%$). This supports our hypothesis that strategy encourages models to learn subtle temporal-spatial variations and improves model generalization ability.
Further, our augmentation strategy exceeds $1.02\%$ than the combination of Mixup and Cutmix, demonstrating the augmentation's effectiveness in TALL for video detection.

\begin{table}[ht]
\centering

\setlength\tabcolsep{4pt}{
\begin{tabular}{cccc}
\hline
Augmentation & Count & CDF & DFDC \\ \hline
None     & -     & 87.60   & 74.32     \\
Cutout & 1        & 89.06    &  74.07  \\
Cutout & 4     & 89.33    & 75.22     \\
Mixup+CutMix      & 4     & 89.75    &  75.33    \\
TALL's mask  & 4     & \textbf{90.79}    & \textbf{76.78}    \\ \hline
\end{tabular}
}
\vspace{0.5em}
\caption{\textbf{Study of the augmentation strategy in TALL.} The count column represents the number of blocks on the thumbnail.}
\label{tb:cutout}
\end{table}

\textbf{Study on window size.} We study the effect of window size on model performance and computational cost. The results are shown in Table~\ref{table:window}. Our window expansion for the first three phases will increase the model performance by 1.74$\%$ AUC.
The results in the second and third rows show that the first three stages of the window getting the largest would not give a boost to the model. Our analysis of a too-large window may weaken the model's ability to learn local information in the sub-image.

\begin{table}[ht]
\centering
\begin{tabular}{ccc}
\hline
Window size   & CDF & DFDC  \\ \hline
(7,7,7,7)     & 85.60   &  73.32     \\
(14,14,14,7)    &\textbf{87.60}  &  \textbf{74.32}    \\
(28,28,28,7) & 86.65    &   74.21     \\
 \hline
\end{tabular}
\vspace{0.5em}
\caption{\textbf{Ablation studies of window size.} All models here are trained without mask augmentation.}
\label{table:window}
\end{table}

\section{Conclusion}

This paper presents a novel perspective on detecting deepfake videos using TALL. TALL is both simple and effective, enabling joint spatio-temporal modeling without any additional costs. TALL representation reveals normal deepfake patterns with local-global contextual features. We further propose a new baseline for deepfake video detection called TALL-Swin, which efficiently captures the inconsistencies between deepfake video frames.
Extensive experiments demonstrate that TALL-Swin achieves promising results for various unseen deepfake types and strong robustness to a wide range of common corruptions.


\section*{Acknowledgment}
This work was partially funded by National Natural Science Foundation of China under Grants (62276256, U21B2045 and U20A20223) and Beijing Nova Program under Grant Z211100002121108.
The authors wish to thank Huaibo Huang and Lijun Sheng in no particular order, for insightful discussions.

{\small
\bibliographystyle{ieee_fullname}
\bibliography{main}

\begin{thebibliography}{10}\itemsep=-1pt

\bibitem{a:2018}
Darius Afchar, Vincent Nozick, Junichi Yamagishi, and Isao Echizen.
\newblock Mesonet: a compact facial video forgery detection network.
\newblock In {\em Proc. WIFS}, pages 1--7, 2018.

\bibitem{vivit}
Anurag Arnab, Mostafa Dehghani, Georg Heigold, Chen Sun, Mario Lucic, and
  Cordelia Schmid.
\newblock Vivit: {A} video vision transformer.
\newblock In {\em Proc. ICCV}, pages 6836--6846, 2021.

\bibitem{Cao_2022_CVPR}
Junyi Cao, Chao Ma, Taiping Yao, Shen Chen, Shouhong Ding, and Xiaokang Yang.
\newblock End-to-end reconstruction-classification learning for face forgery
  detection.
\newblock In {\em Proc. CVPR}, pages 4113--4122, 2022.

\bibitem{carreira2017i3d}
Joao Carreira and Andrew Zisserman.
\newblock Quo vadis, action recognition? a new model and the kinetics dataset.
\newblock In {\em Proc. CVPR}, pages 6299--6308, 2017.

\bibitem{chai2020patch}
Lucy Chai, David Bau, Ser-Nam Lim, and Phillip Isola.
\newblock What makes fake images detectable? understanding properties that
  generalize.
\newblock In {\em Proc. ECCV}, pages 103--120, 2020.

\bibitem{xception}
Fran{\c{c}}ois Chollet.
\newblock Xception: Deep learning with depthwise separable convolutions.
\newblock In {\em Proc. CVPR}, pages 1251--1258, 2017.

\bibitem{deepfake-faceswap}
deepfakes.
\newblock Deepfakes.
\newblock \url{https://github.com/deepfakes/faceswap}.
\newblock 2021-11-13.

\bibitem{devries2017improved}
Terrance DeVries and Graham~W Taylor.
\newblock Improved regularization of convolutional neural networks with cutout.
\newblock {\em arXiv:1708.04552}, 2017.

\bibitem{dfdc}
Brian Dolhansky, Joanna Bitton, Ben Pflaum, Jikuo Lu, Russ Howes, Menglin Wang,
  and Cristian~Canton Ferrer.
\newblock The deepfake detection challenge (dfdc) dataset.
\newblock {\em arXiv:2006.07397}, 2020.

\bibitem{dong2022think}
Chengdong Dong, Ajay Kumar, and Eryun Liu.
\newblock Think twice before detecting gan-generated fake images from their
  spectral domain imprints.
\newblock In {\em Proc. CVPR}, pages 7865--7874, 2022.

\bibitem{dong:2022}
Xiaoyi Dong, Jianmin Bao, Dongdong Chen, Ting Zhang, Weiming Zhang, Nenghai Yu,
  Dong Chen, Fang Wen, and Baining Guo.
\newblock Protecting celebrities from deepfake with identity consistency
  transformer.
\newblock In {\em Proc. CVPR}, pages 9468--9478, 2022.

\bibitem{vit}
Alexey Dosovitskiy, Lucas Beyer, Alexander Kolesnikov, Dirk Weissenborn,
  Xiaohua Zhai, Thomas Unterthiner, Mostafa Dehghani, Matthias Minderer, Georg
  Heigold, Sylvain Gelly, Jakob Uszkoreit, and Neil Houlsby.
\newblock An image is worth 16x16 words: Transformers for image recognition at
  scale.
\newblock In {\em Proc. ICLR}, 2021.

\bibitem{fei2022learning}
Jianwei Fei, Yunshu Dai, Peipeng Yu, Tianrun Shen, Zhihua Xia, and Jian Weng.
\newblock Learning second order local anomaly for general face forgery
  detection.
\newblock In {\em Proc. CVPR}, pages 20270--20280, 2022.

\bibitem{goodfellow2014}
Ian Goodfellow, Jean Pouget-Abadie, Mehdi Mirza, Bing Xu, David Warde-Farley,
  Sherjil Ozair, Aaron Courville, and Yoshua Bengio.
\newblock Generative adversarial networks.
\newblock In {\em Proc. NeurIPS}, volume~27, 2014.

\bibitem{gu2021spatiotemporal}
Zhihao Gu, Yang Chen, Taiping Yao, Shouhong Ding, Jilin Li, Feiyue Huang, and
  Lizhuang Ma.
\newblock Spatiotemporal inconsistency learning for deepfake video detection.
\newblock In {\em Proc. ACM MM}, pages 3473--3481, 2021.

\bibitem{gu2022delving}
Zhihao Gu, Yang Chen, Taiping Yao, Shouhong Ding, Jilin Li, and Lizhuang Ma.
\newblock Delving into the local: Dynamic inconsistency learning for deepfake
  video detection.
\newblock In {\em Proc. AAAI}, pages 744--752, 2022.

\bibitem{haliassos2022realforensics}
Alexandros Haliassos, Rodrigo Mira, Stavros Petridis, and Maja Pantic.
\newblock Leveraging real talking faces via self-supervision for robust forgery
  detection.
\newblock In {\em {Proc. CVPR}}, 2022.

\bibitem{haliassos2021lips}
Alexandros Haliassos, Konstantinos Vougioukas, Stavros Petridis, and Maja
  Pantic.
\newblock Lips don't lie: A generalisable and robust approach to face forgery
  detection.
\newblock In {\em Proc. CVPR}, pages 5039--5049, 2021.

\bibitem{hara2017learning}
Kensho Hara, Hirokatsu Kataoka, and Yutaka Satoh.
\newblock Learning spatio-temporal features with 3d residual networks for
  action recognition.
\newblock In {\em Proc. ICCV Workshops}, pages 3154--3160, 2017.

\bibitem{He_2016_CVPR}
Kaiming He, Xiangyu Zhang, Shaoqing Ren, and Jian Sun.
\newblock Deep residual learning for image recognition.
\newblock In {\em Proc. CVPR}, pages 770--778, 2016.

\bibitem{hsu2022dual}
Gee-Sern Hsu, Chun-Hung Tsai, and Hung-Yi Wu.
\newblock Dual-generator face reenactment.
\newblock In {\em Proc. CVPR}, pages 642--650, 2022.

\bibitem{huang2022fakelocator}
Yihao Huang, Felix Juefei-Xu, Qing Guo, Yang Liu, and Geguang Pu.
\newblock Fakelocator: Robust localization of gan-based face manipulations.
\newblock {\em IEEE Transactions on Information Forensics and Security},
  17:2657--2672, 2022.

\bibitem{ji2022video}
Ge-Peng Ji, Guobao Xiao, Yu-Cheng Chou, Deng-Ping Fan, Kai Zhao, Geng Chen, and
  Luc Van~Gool.
\newblock Video polyp segmentation: A deep learning perspective.
\newblock {\em MIR}, 19(6):531--549, 2022.

\bibitem{ji2023masked}
Ge-Peng Ji, Mingchen Zhuge, Dehong Gao, Deng-Ping Fan, Christos Sakaridis, and
  Luc~Van Gool.
\newblock Masked vision-language transformer in fashion.
\newblock {\em MIR}, 20(3):421--434, 2023.

\bibitem{jia2021inconsistency}
Gengyun Jia, Meisong Zheng, Chuanrui Hu, Xin Ma, Yuting Xu, Luoqi Liu, Yafeng
  Deng, and Ran He.
\newblock Inconsistency-aware wavelet dual-branch network for face forgery
  detection.
\newblock {\em IEEE Transactions on Biometrics, Behavior, and Identity
  Science}, 3(3), 2021.

\bibitem{jiang2020deeperforensics}
Liming Jiang, Ren Li, Wayne Wu, Chen Qian, and Chen~Change Loy.
\newblock Deeperforensics-1.0: A large-scale dataset for real-world face
  forgery detection.
\newblock In {\em Proc. CVPR}, pages 2889--2898, 2020.

\bibitem{karras2019style}
Tero Karras, Samuli Laine, and Timo Aila.
\newblock A style-based generator architecture for generative adversarial
  networks.
\newblock In {\em Proc. CVPR}, pages 4401--4410, 2019.

\bibitem{khan:2021}
Sohail~Ahmed Khan and Hang Dai.
\newblock Video transformer for deepfake detection with incremental learning.
\newblock In {\em {Proc. ACM MM}}, pages 1821--1828, 2021.

\bibitem{khormali2022dfdt}
Aminollah Khormali and Jiann-Shiun Yuan.
\newblock Dfdt: An end-to-end deepfake detection framework using vision
  transformer.
\newblock {\em Applied Sciences}, page 2953, 2022.

\bibitem{kingma2014adam}
Diederik~P Kingma and Jimmy Ba.
\newblock Adam: A method for stochastic optimization.
\newblock {\em arXiv:1412.6980}, 2014.

\bibitem{li2021fdfl}
Jiaming Li, Hongtao Xie, Jiahong Li, Zhongyuan Wang, and Yongdong Zhang.
\newblock Frequency-aware discriminative feature learning supervised by
  single-center loss for face forgery detection.
\newblock In {\em Proc. CVPR}, pages 6458--6467, 2021.

\bibitem{FaceXray}
Lingzhi Li, Jianmin Bao, Ting Zhang, Hao Yang, Dong Chen, Fang Wen, and Baining
  Guo.
\newblock Face x-ray for more general face forgery detection.
\newblock In {\em Proc. CVPR}, pages 5001--5010, 2020.

\bibitem{li2019DSP}
Yuezun Li and Siwei Lyu.
\newblock Exposing deepfake videos by detecting face warping artifacts.
\newblock In {\em {Proc. CVPR Workshops}}, pages 656--663, 2019.

\bibitem{li2020celeb}
Yuezun Li, Xin Yang, Pu Sun, Honggang Qi, and Siwei Lyu.
\newblock Celeb-df: A large-scale challenging dataset for deepfake forensics.
\newblock In {\em Proc. CVPR}, pages 3207--3216, 2020.

\bibitem{Liu_2021_CVPR}
Honggu Liu, Xiaodan Li, Wenbo Zhou, Yuefeng Chen, Yuan He, Hui Xue, Weiming
  Zhang, and Nenghai Yu.
\newblock Spatial-phase shallow learning: Rethinking face forgery detection in
  frequency domain.
\newblock In {\em Proc. CVPR}, pages 772--781, 2021.

\bibitem{liu2021swin}
Ze Liu, Yutong Lin, Yue Cao, Han Hu, Yixuan Wei, Zheng Zhang, Stephen Lin, and
  Baining Guo.
\newblock Swin transformer: Hierarchical vision transformer using shifted
  windows.
\newblock In {\em Proc. ICCV}, pages 10012--10022, 2021.

\bibitem{liu2020global}
Zhengzhe Liu, Xiaojuan Qi, and Philip~HS Torr.
\newblock Global texture enhancement for fake face detection in the wild.
\newblock In {\em Proc. CVPR}, pages 8060--8069, 2020.

\bibitem{faceswap}
MarekKowalski.
\newblock Faceswap.
\newblock \url{https://github.com/MarekKowalski/FaceSwap/}.
\newblock 2021-11-13.

\bibitem{masi2020two}
Iacopo Masi, Aditya Killekar, Royston~Marian Mascarenhas, Shenoy~Pratik
  Gurudatt, and Wael AbdAlmageed.
\newblock Two-branch recurrent network for isolating deepfakes in videos.
\newblock In {\em Proc. ECCV}, pages 667--684, 2020.

\bibitem{mirsky2021creation}
Yisroel Mirsky and Wenke Lee.
\newblock The creation and detection of deepfakes: A survey.
\newblock {\em ACM CSUR}, 54(1):1--41, 2021.

\bibitem{neimark2021video}
Daniel Neimark, Omri Bar, Maya Zohar, and Dotan Asselmann.
\newblock Video transformer network.
\newblock In {\em Proc. ICCV}, pages 3163--3172, 2021.

\bibitem{nirkin2021deepfake}
Yuval Nirkin, Lior Wolf, Yosi Keller, and Tal Hassner.
\newblock Deepfake detection based on discrepancies between faces and their
  context.
\newblock {\em IEEE Transactions on Pattern Analysis and Machine Intelligence},
  44(10):6111--6121, 2021.

\bibitem{qian2020}
Yuyang Qian, Guojun Yin, Lu Sheng, Zixuan Chen, and Jing Shao.
\newblock Thinking in frequency: Face forgery detection by mining
  frequency-aware clues.
\newblock In {\em Proc. ECCV}, pages 86--103, 2020.

\bibitem{fan-iclr2022}
Rameswar~Panda Quanfu~Fan, Richard~Chen.
\newblock Can an image classifier suffice for action recognition?
\newblock In {\em ICLR}, 2022.

\bibitem{FaceForensics}
Andreas Rossler, Davide Cozzolino, Luisa Verdoliva, Christian Riess, Justus
  Thies, and Matthias Niessner.
\newblock Faceforensics++: Learning to detect manipulated facial images.
\newblock In {\em Proc. ICCV}, pages 1--11, 2019.

\bibitem{sabir2019gru}
Ekraam Sabir, Jiaxin Cheng, Ayush Jaiswal, Wael AbdAlmageed, Iacopo Masi, and
  Prem Natarajan.
\newblock Recurrent convolutional strategies for face manipulation detection in
  videos.
\newblock In {\em {Proc. CVPR Workshops}}, pages 80--87, 2019.

\bibitem{Selvaraju_2017_ICCV}
Ramprasaath~R. Selvaraju, Michael Cogswell, Abhishek Das, Ramakrishna Vedantam,
  Devi Parikh, and Dhruv Batra.
\newblock Grad-cam: Visual explanations from deep networks via gradient-based
  localization.
\newblock In {\em Proc. ICCV}, pages 618--626, 2017.

\bibitem{shi2022semanticstylegan}
Yichun Shi, Xiao Yang, Yangyue Wan, and Xiaohui Shen.
\newblock Semanticstylegan: Learning compositional generative priors for
  controllable image synthesis and editing.
\newblock In {\em Proc. CVPR}, pages 11254--11264, 2022.

\bibitem{sun2022fenerf}
Jingxiang Sun, Xuan Wang, Yong Zhang, Xiaoyu Li, Qi Zhang, Yebin Liu, and Jue
  Wang.
\newblock Fenerf: Face editing in neural radiance fields.
\newblock In {\em Proc. CVPR}, pages 7672--7682, 2022.

\bibitem{tan2019efficientnet}
Mingxing Tan and Quoc Le.
\newblock Efficientnet: Rethinking model scaling for convolutional neural
  networks.
\newblock In {\em Proc. ICML}, pages 6105--6114, 2019.

\bibitem{Thies:2019}
Justus Thies, Michael Zollh{\"{o}}fer, and Matthias Nie{\ss}ner.
\newblock Deferred neural rendering: Image synthesis using neural textures.
\newblock {\em ACM TOG}, 38(4):1--12, 2019.

\bibitem{thies2016face2face}
Justus Thies, Michael Zollhofer, Marc Stamminger, Christian Theobalt, and
  Matthias Nie{\ss}ner.
\newblock Face2face: Real-time face capture and reenactment of rgb videos.
\newblock In {\em Proc. ICCV}, pages 2387--2395, 2016.

\bibitem{verdoliva2020media}
Luisa Verdoliva.
\newblock Media forensics and deepfakes: an overview.
\newblock {\em IEEE Journal of Selected Topics in Signal Processing},
  14(5):910--932, 2020.

\bibitem{wang2021representative}
Chengrui Wang and Weihong Deng.
\newblock Representative forgery mining for fake face detection.
\newblock In {\em Proc. CVPR}, pages 14923--14932, 2021.

\bibitem{wang:2022}
Junke Wang, Zuxuan Wu, Wenhao Ouyang, Xintong Han, Jingjing Chen, Yu-Gang
  Jiang, and Ser-Nam Li.
\newblock M2tr: Multi-modal multi-scale transformers for deepfake detection.
\newblock In {\em Proc. ICMR}, pages 615--623, 2022.

\bibitem{wang2022adt}
Ping Wang, Kunlin Liu, Wenbo Zhou, Hang Zhou, Honggu Liu, Weiming Zhang, and
  Nenghai Yu.
\newblock Adt: Anti-deepfake transformer.
\newblock In {\em Proc. ICASSP}, pages 2899--1903, 2022.

\bibitem{wang2020cnnaug}
Sheng-Yu Wang, Oliver Wang, Richard Zhang, Andrew Owens, and Alexei~A Efros.
\newblock Cnn-generated images are surprisingly easy to spot... for now.
\newblock In {\em Proc. CVPR}, 2020.

\bibitem{wodajo:2021}
Deressa Wodajo and Solomon Atnafu.
\newblock Deepfake video detection using convolutional vision transformer.
\newblock {\em arXiv:2102.11126}, 2021.

\bibitem{Yang_masked}
Ziming Yang, Jian Liang, Yuting Xu, Xiao-Yu Zhang, and Ran He.
\newblock Masked relation learning for deepfake detection.
\newblock {\em IEEE Transactions on Information Forensics and Security}, pages
  1696--1708, 2023.

\bibitem{yun2019cutmix}
Sangdoo Yun, Dongyoon Han, Seong~Joon Oh, Sanghyuk Chun, Junsuk Choe, and
  Youngjoon Yoo.
\newblock Cutmix: Regularization strategy to train strong classifiers with
  localizable features.
\newblock In {\em Proc. ICCV}, pages 6023--6032, 2019.

\bibitem{zeng2022sketchedit}
Yu Zeng, Zhe Lin, and Vishal~M Patel.
\newblock Sketchedit: Mask-free local image manipulation with partial sketches.
\newblock In {\em Proc. CVPR}, pages 5951--5961, 2022.

\bibitem{zhang2017mixup}
Hongyi Zhang, Moustapha Cisse, Yann~N Dauphin, and David Lopez-Paz.
\newblock Mixup: Beyond empirical risk minimization.
\newblock {\em Proc. ICLR}, 2018.

\bibitem{zhang2021vidtr}
Yanyi Zhang, Xinyu Li, Chunhui Liu, Bing Shuai, Yi Zhu, Biagio Brattoli, Hao
  Chen, Ivan Marsic, and Joseph Tighe.
\newblock Vidtr: Video transformer without convolutions.
\newblock In {\em Proc. ICCV}, pages 13577--13587, 2021.

\bibitem{zhao2023istvt}
Cairong Zhao, Chutian Wang, Guosheng Hu, Haonan Chen, Chun Liu, and Jinhui
  Tang.
\newblock Istvt: Interpretable spatial-temporal video transformer for deepfake
  detection.
\newblock {\em IEEE Transactions on Information Forensics and Security},
  18:1335--1348, 2023.

\bibitem{zhao2021multi}
Hanqing Zhao, Wenbo Zhou, Dongdong Chen, Tianyi Wei, Weiming Zhang, and Nenghai
  Yu.
\newblock Multi-attentional deepfake detection.
\newblock In {\em Proc. CVPR}, pages 2185--2194, 2021.

\bibitem{zhao:2022}
Hanqing Zhao, Wenbo Zhou, Dongdong Chen, Weiming Zhang, and Nenghai Yu.
\newblock Self-supervised transformer for deepfake detection.
\newblock {\em arXiv:2203.01265}, 2022.

\bibitem{self_cons}
Tianchen Zhao, Xiang Xu, Mingze Xu, Hui Ding, Yuanjun Xiong, and Wei Xia.
\newblock Learning self-consistency for deepfake detection.
\newblock In {\em Proc. CVPR}, pages 15023--15033, 2021.

\bibitem{zheng:2021}
Yinglin Zheng, Jianmin Bao, Dong Chen, Ming Zeng, and Fang Wen.
\newblock Exploring temporal coherence for more general video face forgery
  detection.
\newblock In {\em Proc. ICCV}, pages 15044--15054, 2021.

\bibitem{zhu2021face}
Xiangyu Zhu, Hao Wang, Hongyan Fei, Zhen Lei, and Stan~Z Li.
\newblock Face forgery detection by 3d decomposition.
\newblock In {\em Proc. CVPR}, pages 2929--2939, 2021.

\bibitem{zi2020add}
Bojia Zi, Minghao Chang, Jingjing Chen, Xingjun Ma, and Yu-Gang Jiang.
\newblock Wilddeepfake: A challenging real-world dataset for deepfake
  detection.
\newblock In {\em {Proc. ACM MM}}, pages 2382--2390, 2020.

\end{thebibliography}
}





\end{document}